\documentclass[orivec]{llncs}

\usepackage{mathtools}

\usepackage{wrapfig}
\usepackage{stmaryrd}
\usepackage{xspace}
\usepackage{units}
\usepackage{xcolor}

\usepackage{amsfonts}
\usepackage{amsmath}
\usepackage{amssymb}
\usepackage{mathtools}

\usepackage{tabularx}
\usepackage{subfig}

\usepackage{graphicx}
\usepackage[export]{adjustbox}

\usepackage{gensymb}

\usepackage{multicol}
\usepackage{colortbl}

\usepackage{listings}

\usepackage[defblank]{paralist}
\usepackage{booktabs}

\usepackage{comment}

\usepackage[T1]{fontenc}

  \usepackage[textsize=tiny,colorinlistoftodos]{todonotes}

\newcommand{\conj}{\ensuremath{{\sf conj}}\xspace}

\newcommand{\cert}{\ensuremath{{\sf cert}}\xspace}

\newcommand{\sql}{\ensuremath{{\sf sql}}\xspace}

\newcommand{\set}[1]{\{ #1 \}\xspace}

\newcommand{\incl}{\subseteq\xspace}

\newcommand{\dllite}{\ensuremath{\textit{DL-Lite}}\xspace}
\newcommand{\dlliteA}{\ensuremath{\textit{DL-Lite}_{\cal A}}\xspace}

\newcommand{\dlliteAAggr}{\ensuremath{\textit{DL-Lite}_{\cal A}^{\sf agg}}\xspace}

\newcommand{\bagl}{\{\hspace{-.5ex}|\xspace}
\newcommand{\bagr}{|\hspace{-.5ex}\}\xspace}

\newcommand{\bag}[1]{\bagl #1 \bagr \xspace}

\newcommand{\agg}{{\sf{agg}}\xspace}
\renewcommand{\min}{{\sf{min}}\xspace}
\newcommand{\minf}{{\sf{min}}\xspace}
\renewcommand{\max}{{\sf{max}}\xspace}
\newcommand{\maxf}{{\sf{max}}\xspace}
\newcommand{\avg}{{\sf{avg}}\xspace}
\newcommand{\avgf}{{\sf{avg}}\xspace}
\newcommand{\countf}{{\sf{count}}\xspace}
\newcommand{\countdf}{{\sf{countd}}\xspace}
\newcommand{\sumf}{{\sf{sum}}\xspace}

\newcommand{\owlaql}{\dlliteAAggr}

\newcommand{\isa}{\sqsubseteq\xspace}

\renewcommand{\O}{\ensuremath{\mathcal{O}}\xspace}
\newcommand{\D}{\ensuremath{\mathcal{D}}\xspace}
\newcommand{\M}{\ensuremath{\mathcal{M}}\xspace}
\newcommand{\I}{\ensuremath{\mathcal{I}}\xspace}
\newcommand{\Q}{\ensuremath{\mathcal{Q}}\xspace}

\definecolor{darkred}{rgb}{0.7,0.1,0.1}

\newcommand{\starql}{{\scshape STARQL}\xspace}

\def\exareme{{\scshape Exareme}\xspace}
\def\exastream{{\scshape ExaStream}\xspace}
\def\dsms{\emph{DSMS}\xspace}

\def\udf{\emph{UDF}\xspace}

\def\obda{\emph{OBDA}\xspace}

\newcommand{\ignore}[1]{}

\def\mws{\emph{MWS}\xspace}

\def\exareme{{\scshape Exareme}\xspace}
\def\exastream{{\scshape ExaStream}\xspace}
\def\dsms{\emph{DSMS}\xspace}

\def\udf{\emph{UDF}\xspace}

\def\obda{\emph{OBDA}\xspace}

\newcommand\sqls{{\scshape SQL\ensuremath{^{\varoplus}}}\xspace}
\def\starqlcql{{\scshape STARQL2}\sqls\xspace}

\title{
Towards Analytics Aware Ontology Based Access \\ to Static and Streaming Data\\
(Extended Version)%
\thanks{This work was partially funded by the EU project
Optique (FP7-ICT-318338) and
the EPSRC projects MaSI$^3$, DBOnto, and ED$^3$
}
}

\newcommand{\authspace}{\hspace{.6ex}}
\author{%
\mbox{}\hspace{-3ex}
E. Kharlamov$^{1}$%
\authspace%
Y. Kotidis$^{2}$\authspace%
T. Mailis$^{3}$\authspace%
C. Neuenstadt$^{4}$\authspace%
C. Nikolaou$^{1}$\authspace%
\"O. \"Oz\c{c}ep$^{4}$\authspace\\%
C. Svingos$^{3}$\authspace
D. Zheleznyakov$^{1}$\authspace
S. Brandt$^{5}$\authspace
I. Horrocks$^{1}$\authspace%
Y. Ioannidis$^{3}$\authspace\\%
S. Lamparter$^{5}$\authspace
R. M\"oller$^{4}$
}		
\institute{
\mbox{}\hspace{-4ex}
	$^1$University of Oxford\authspace\authspace\authspace\authspace
    $^2$Athens University of Economics and Business\authspace\authspace\authspace\authspace\\
	$^3$University of Athens\authspace\authspace\authspace\authspace
	$^4$University of L\"ubeck\authspace\authspace\authspace\authspace
	$^5$Siemens Corporate Technology
}

\begin{document}

\maketitle

\begin{abstract}
Real-time analytics that requires integration and aggregation of heterogeneous and distributed streaming and static data is a typical task in many industrial scenarios such as diagnostics of turbines in Siemens. OBDA approach has a great potential to facilitate such tasks; however, it has a number of limitations in dealing with analytics that restrict its use in important industrial applications. Based on our experience with Siemens, we argue that in order to overcome those limitations OBDA should be extended and become analytics, source, and cost aware. In this work we propose such an extension. In particular, we propose an ontology, mapping, and query language for OBDA, where aggregate and other analytical functions are first class citizens. Moreover, we develop query optimisation techniques that allow to efficiently process analytical tasks over static and streaming data. We implement our approach in a system and evaluate our system with Siemens turbine data.
\end{abstract}

%
%

%


\section{Introduction}
\label{sec:introduction}

\emph{Ontology Based Data Access (OBDA)}~\cite{Calvanese:2009tf}
	is an approach to access
	information stored in multiple datasources 
	via an abstraction layer
	that mediates between 
	the datasources and data consumers.
This layer uses an \emph{ontology} to
	provide a uniform conceptual schema that describes 
	the problem domain of the underlying data 
	independently of how and where the data is stored,
	and declarative \emph{mappings} 
	to specify how the ontology is related to the data
	by relating elements of the ontology
	to queries over datasources.
The ontology and mappings are used to \emph{transform} 
	queries over ontologies, i.e., \emph{ontological queries},
	into \emph{data queries} over datasources.
As well as abstracting away
from details of data storage and access, 
the ontology and mappings provide a declarative, modular and query-independent
specification of both the conceptual model and its relationship to the data sources;
this simplifies development and maintenance and allows for
easy integration with existing data management infrastructure.
%

A number of systems that at least partially 
	implement OBDA
	have been recently developed;
	they include
	D2RQ~\cite{Bizer:2004ug}, 
	Mastro~\cite{Calvanese:2011dl}, 
	morph-RDB~\cite{Priyatna:2014fw}, 
	Ontop~\cite{RodriguezMuro:2013cs}, 
	OntoQF~\cite{Munir:2012ik},
	Ultrawrap~\cite{Sequeda:2013ij},
	Virtuoso, 
	Spyder, 
	and others~\cite{DBLP:conf/semweb/CalbimonteCG10,DBLP:conf/semweb/FischerSB13a}.
Some of them were successfully used in various
	applications 
	including cultural 
	heritage~\cite{DBLP:journals/eaai/CalvaneseLMRRR16},
	governmental 
	organisations~\cite{DBLP:journals/pvldb/CiviliCGLLLMPRRSS13}, 
	and 
	industry~\cite{statoil2015,DBLP:conf/semweb/KharlamovSOZHLRSW14}.
%
Despite their success,
	OBDA systems, however, are not tailored towards 
	analytical tasks that are naturally 
	based on data aggregation and correlation.
Moreover, they offer a limited or no support 
	for queries that combine streaming and static data.
A typical scenario that requires both analytics 
	and access to static and streaming data 
	is diagnostics and monitoring of turbines in Siemens.

Siemens has several service centres 
	dedicated to diagnostics of 
	thousands of power-generation appliances located across the 
	globe~\cite{DBLP:conf/semweb/KharlamovSOZHLRSW14}.
One typical task of such a centre is
	to detect in real-time 
	potential faults of a turbine 
	caused by, e.g., 
	an undesirable pattern in temperature's behaviour
	within various components of the turbine.
Consider a (simplified) example of such a task:
\begin{quote}
\emph{
In a given turbine 
	report all temperature sensors that
	are reliable,
	i.e., 
	with the average score of validation tests at least 90\%,
	and 
	whose measurements within the last 10 min 
	were similar,
	i.e., Pearson correlated by at least 0.75,
	to measurements reported last year by a reference sensor
	that had been functioning in a critical mode.
}
\end{quote}
This task requires 
	to extract, aggregate, and correlate 
	static data about the turbine's structure,
	streaming data produced by
	up to 2,000 sensors installed in different parts of the turbine,
	and historical operational data of the reference sensor 
	stored in multiple datasources.
Accomplishing such a task 
	currently requires 
	to pose a collection of hundreds of queries,
	the majority of which are 
	semantically the same (they ask about temperature),
	but syntactically differ
	(they are over different schemata).
Formulating and executing so many queries 
	and then assembling the computed answers
	take up to 80\% of the overall diagnostic 
	time that Siemens engineers typically have to 
	spend~\cite{DBLP:conf/semweb/KharlamovSOZHLRSW14}.
The use of ODBA, however, would allow to save 
	a lot of this time since 
	ontologies
	can help to `hide' 
	the technical details of \emph{how} the data is 
	produced, represented, and stored in data sources,
	and to show only \emph{what} this data is about.
Thus, 
	one would be able to formulate this diagnostic task 
	using only one ontological query instead of 
	a collection of hundreds data queries that
	today have to be written or configured by IT specialists.
Clearly, this collection of queries does not disappear:
	the OBDA query tranformation
	will automatically 
	compute them from the the high-level ontological query
	using the ontology and mappings.

Siemens analytical tasks 
	as the one in the example scenario typically
	make heavy use of aggregation and correlation functions 
	as well as arithmetic operations.
	In our 	running example, the aggregation function $\min$ 
	and the comparison operator $\geq$
 	are used 
	to specify what
	makes a sensor reliable and to define a threshold for similarity. 
Performing such operations 
	only in ontological queries, 
	or only in data queries specified in the mappings
	is not satisfactory.
In the case of ontological queries, 
	all relevant values should be retrieved prior 
	to performing grouping and arithmetic operations. 
This can be highly inefficient, as it fails
	to exploit source capabilities (e.g., access to pre-computed averages), 
	and value retrieval may be slow and/or costly, e.g., 
	when relevant values are stored remotely.
	Moreover, it adds to the complexity of application queries, and thus limits the benefits of
	the abstraction layer.
In the case of source queries, 
	aggregation functions and comparison operators may be used in mapping queries.
This is brittle and inflexible,
	as values such as 90\% and 0.75, which are used to define
	`reliable sensor' and `similarity', 
	cannot be specified in
	the ontological query, but must be `hard-wired' in the mappings,
	unless an appropriate extension to the query language or the ontology
	are developed.
In order to address these issues, 
	OBDA should become
\begin{quote}
	\emph{analytics-aware} by supporting
	declarative representations of basic analytics operations and using
	these to efficiently answer higher level queries.
\end{quote}
In practice this requires
	enhancing
	OBDA technology with ontologies, mappings, 
	and query languages capable  of capturing
	operations used in analytics, 
	but also extensive modification of OBDA
	query preprocessing components, 
	i.e., reasoning and query transformation, to support these enhanced languages.


Moreover, 
	analytical tasks as in the example scenario 
	should typically be executed
	continuously
	in data intensive 
	and highly distributed environments 
	of streaming and static data.
Efficiency of such execution requires non-trivial query optimisation.
However, optimisations in existing OBDA systems 
	are usually limited to minimisation 
	of the textual size of the generated
	queries, e.g.~\cite{RodrguezMuro:2012tb}, 
	with little support for distributed query processing,
	and no support for optimisation
	for continuous queries over sequences of numerical data
	and, in particular, 
	computation of data correlation and aggregation 
	across static and streaming data.
In order to address these issues, 
	OBDA should become
\begin{quote}
	\emph{source and cost aware} by supporting
	both static and streaming data sources
	and offering a  
	robust  query planning component and indexing 
	that can estimate the cost of different plans,
	and use such estimates to produce low-cost plans.
\end{quote}
Note that the existence of materialised and pre-computed 
	subqueries relevant to analytics within sources
	and archived historical data that  should be correlated with 
	current streaming data 
	 implies that there is a range of query plans which can
	 differ dramatically with respect to data transfer
	 and query execution time.

In this paper we make the first step 
to extend OBDA systems towards becoming analytics,
source, and cost aware and thus 
meeting Siemens requirements for turbine diagnostics tasks.
In particular, our contributions are the following:
\begin{itemize} 
	\item 
		We proposed
		analytics-aware OBDA components, i.e.,
		\begin{inparaenum}[\it (i)]
			\item 		
				ontology language \owlaql that 
				extends \dlliteA with 
				aggregate functions as first class citizens,
			\item 
				query language \starql over 
				ontologies that combine streaming 
				and static data, and
			\item 
				a mapping language
				relating \owlaql vocabulary 
				and \starql constructs 
				with 
				relational queries over static and 
				streaming data.
		\end{inparaenum}
	\item 
		We developed 
		efficient query transformation techniques that allow to
		turn \starql queries over \owlaql ontologies,
		into data queries using our mappings.
	\item 
		We developed source and cost aware 
		\begin{inparaenum}[\it (i)]
			\item 		
			optimisation techniques
			for processing complex analytics on both 
			static and streaming data, 
			including adaptive indexing schemes and
			pre-computation of frequent aggregates 
			on user queries,
			and
			\item 
			elastic infrastructure 
			that automatically distributes analytical
			computations and data over a computational cloud
			for fastet query execution. 
		\end{inparaenum}		
	\item 
		We implemented
		\begin{inparaenum}[\it (i)]
			\item 		
			a highly optimised engine \exastream capable 
			of handling complex streaming and static queries in real time,
			\item 
			a dedicated \starqlcql translator that transforms 
			\starql queries into queries
			over static and streaming data,
			\item 
			an integrated OBDA system
			that relies on our and third party components.
		\end{inparaenum}				
	\item 
		We conducted
		a performance evaluation of  our OBDA system with 
		large scale Siemens simulated data
		using analytical tasks.
\end{itemize}

\noindent The paper is organised as follows:
Sec.~\ref{sec:analytics-aware-obda}
presents our analytics-aware ontology, query,
and mapping languages as well as
query optimisation techniques;
Sec.~\ref{sec:implementation} and~\ref{sec:evaluation}
discuss implementation of our system
and	presents experiments;
Sec.~\ref{sec:related-work} discusses related work.

\section{Analytics Aware OBDA for Static and Streaming Data}
\label{sec:analytics-aware-obda}
In this section we first introduce 
	our analytics-aware ontology language 
	\owlaql
	(Sec.~\ref{sec:ontology-language})
 	for capturing static aspects
 	of the domain of interest.
In \owlaql ontologies, aggregate functions are 
	treated as first class citizens. 
Then, in Sec~\ref{sec:query-language}
	we will introduce a query language \starql
	that allows to combine 
	static conjunctive queries 
	over \owlaql
	with continuous diagnostic queries
	that involve 
	simple combinations of time aware data attributes,
	time windows, 
	and functions, e.g., correlations
	over streams of attribute values.
Using \starql queries one can retrieve
	entities, e.g., sensors, 
	that pass two `filters':
	static and continuous.
In our running example 
	a static `filter' checks whether a sensor is reliable,
	while a continuous `filter' checks whether
	the measurements of the sensor
	are Pearson correlated with
	the measurements of reference sensor.
In Sec.~\ref{sec:mapping-language}
	we will explain how to 
	translate \starql queries into data queries 
	by mapping \owlaql concepts, properties, and attributes 
	occurring in queries to database schemata
	and 
	by mapping functions and constructs 
	of \starql continuous `filters' 
	into corresponding functions and constructs
	over databases.
Finally,
	in Sec.~\ref{sec:query-optimisation}
	we discuss how to optimise resulting data queries.

\subsection{Ontology Language}
\label{sec:ontology-language}

Our ontology language, $\dlliteAAggr$,
	is an extension of
	$\dlliteA$~\cite{Calvanese:2009tf}
	with concepts that are based on aggregation 
	of attribute values.
The semantics for such concepts adapts
	the closed-world semantics~\cite{Lutz:2012vh}.
The main reason why we rely on this semantics is
	to avoid the problem of 
	empty answers for aggregate queries 
	under the certain answers
	semantics~\cite{Calvanese:2008hp,Kostylev:2015dx}.
In $\dlliteAAggr$ we distinguish between individuals
	and data values
	from countable sets $\Delta$ and $D$
	that intuitively correspond
	to the datatypes of RDF.
We also distinguish
	between atomic roles $P$ 
	that denote binary 
	relations between pairs of individuals,	
	and 
	attributes $F$
	that denote binary relations 
	between individuals and data values.	
For simplicity of presentation we assume 
	that $D$ is the set of rational numbers.
Let $\agg$ be an aggregate function,
	e.g., 
	$\minf$, $\maxf$, $\countf$, 
	$\countdf$, $\sumf$, or 
	$\avgf$, 
	and
	let 
	$\circ$ be a comparison predicate on 
	rational numbers, 
	e.g., $\geq, \leq, <, >, =, $ or $ \neq$.

\subsubsection{$\dlliteAAggr$ Syntax.}	
The grammar for concepts and roles in $\dlliteAAggr$
	is as follows:
	\begin{align*}
		B \rightarrow A \mid \exists R,
		\quad
		C \rightarrow B \mid \exists F,
		\quad
		E \rightarrow \circ_r(\agg\ F),
		\quad
		R \rightarrow P \mid P^-,
	\end{align*}
where $F$, $P$, $\agg$, and $\circ$ are as above,
		$r$ is a rational number,
		$A$, $B$, $C$ and $E$ are atomic, basic, extended 
		and aggregate concepts, respectively,
		and 
		$R$ is a basic role.

A $\dlliteAAggr$ ontology \O 
	is a finite set of 
	axioms.
We consider two types of axioms:
	\emph{aggregate} axioms of the form $E \isa B$ and
	\emph{regular} axioms that take one of the following forms:
	\begin{inparaenum}[\it (i)]
	\item 	
	\emph{inclusions} of the form
	$C \isa B$,
	$R_1 \isa R_2$, and
	$F_1 \isa F_2$,
	\item 
	\emph{functionality} axioms 
	$({\sf funct}\ R)$ 
	and $({\sf funct}\ F),$		
	\item 
	or \emph{denials} of the form
	$B_1 \sqcap B_2 \sqsubseteq \bot$,
	$R_1 \sqcap R_2 \sqsubseteq \bot$, and
	$F_1 \sqcap F_2 \sqsubseteq \bot$.
	\end{inparaenum}
As in \dlliteA, a $\dlliteAAggr$ \emph{dataset} $\D$
	is a finite set of assertions of the form:
	$A(a)$, $R(a,b)$, and $F(a,v)$.

We require that 
	if $({\sf funct}\ R)$ (resp., $({\sf funct}\ F)$) is in $\O$, 
	then $R' \isa R$ (resp., $F' \isa F$) is \emph{not} in $\O$ 
	for any $R'$ (resp., $F'$).
This syntactic condition, as wel as the fact that 
	we do not allow concepts of the form $\exists F$
	and aggregate concepts
	to appear on the right-hand side of inclusions
	ensure good computational properties of \dlliteAAggr.
The former is inherited from \dlliteA,
	while the latter can be shown using techniques of~\cite{Lutz:2012vh}.


Consider the ontology capturing
	the reliability of sensors 
	as in our running example:
\begin{align}	
	\label{eq:ontology}
	\mathit{precisionScore} \sqsubseteq \mathit{testScore},
	\quad
	\geq_{0.9} (\minf\  \mathit{testScore}) \sqsubseteq \sf 	
	\mathit{Reliable},
\end{align}
where $\mathit{Reliable}$ is a concept,
	$precisionScore$ and
	$\mathit{testScore}$ are attributes,
	and finally 
	$\geq_{0.9}\sf (min\ \mathit{testScore})$ 
	is an aggregate concept that captures individuals 
	with one or more $\mathit{testScore}$ values 
	whose minimum is at least $0.9$.

\subsubsection{$\dlliteAAggr$ Semantics.}	
We define the semantics of \dlliteAAggr 
	in terms of first-order
	interpretations 
	over the union of the countable domains $\Delta$ and $D$.
We assume the unique name assumption and that constants are interpreted as themselves,
	i.e., $a^\I = a$ for each constant $a$;
	moreover, interpretations of 
	regular concepts, roles, and attributes
	are defined as usual 
	(see~\cite{Calvanese:2009tf} for details) and  for 
	aggregate concepts as follows: 
	\[  
	(\circ_r (\agg\ F))^\I
	=
	\set{a \in \Delta 
			\mid
			\agg\bag{v \in D \mid (a,v) \in F^\I}
			\circ r
			}.
	\]	 
Here $\bag{\cdot}$ denotes a multi-set.
Similarly to~\cite{Lutz:2012vh},
	we say that an interpretation 
	\I is a \emph{model} of $\O\cup\D$
	if two conditions hold:
	\begin{inparaenum}[\it (i)]
		\item 
		$\I \models \O \cup \D$, i.e., \I is a first-order model of $\O \cup \D$ 
		and
		\item  
		$F^\I=\set{(a,v) \mid F(a,v) \text{ is in the 
		deductive closure of } \D \text{ with } \O}$
		for each attribute $F$.
Here, by deductive closure of $\D$ with $\O$ we assume a dataset
	that can be obtained from $\D$ using the chasing procedure with $\O$,
	as described in~\cite{Calvanese:2009tf}.
		%
	\end{inparaenum}
One can show that for \dlliteAAggr 
satisfiability of $\O\cup\D$ can be checked in 
time polynomial in $|\O\cup\D|$.

As an example consider a dataset consisting of assertions:
$\mathit{precisionScore}(s_1,0.9)$,
$\mathit{testScore}(s_2,0.95)$, and
$\mathit{testScore}(s_3,0.5)$.
Then, for every model $\I$ of these assertions and the axioms 
in Eq.~\eqref{eq:ontology}, it holds that
$(\geq_{0.9}(\min\ \mathit{precisionScore}))^\I =	\set{s_1}$,
$(\geq_{0.9}(\min\ \mathit{testScore}))^\I =	\set{s_1, s_2}$, and
thus
$\set{s_1, s_2} \incl \mathit{Reliable}^\I$.

\subsubsection{Query Answering.}
Let \Q be the class of conjunctive queries 
	over concepts, roles, and attributes,
	i.e., each query $q\in\Q$ is an expression of the form:
\(
	q(\vec x) \text{ :- }
	\conj(\vec x),
\)
where $q$ is of arity $k$, $\conj$ is a conjunction of 
atoms $A(u)$, $E(v)$, $R(w,z)$, or $F(w,z)$,
and $u$, $v$, $w$, $z$ are from $\vec x$.
Following the standard approach for ontologies,
we adapt certain answers semantics for query answering:
\[\cert(q, \O, \D) = 
	\set{ \vec t \in (\Delta \cup D)^k \mid 
	\I \models \conj(\vec t)
	\text{ for each model } \I \text{ of } \O\cup\D
	}.
\]
Continuing with our example,
	consider the query:
		$q(x) \text{ :- } \mathit{Reliable}(x)$
 that asks for
	reliable sensors.
The set of certain answers $\cert(q, \O, \D)$ for this $q$ over
	the example ontology and dataset
	is $\set{s_1, s_2}$.

By relying on Theorem~1 of~\cite{Lutz:2012vh}
	and the fact that
	each aggregate concept behaves like a
	\dllite closed predicate of~\cite{Lutz:2012vh},
	in the sense that its 
	interpretation---given an ontology \O and dataset \D---is 
	determined and fixed by \D,
 	one can show that conjunctive query answering 
	in \dlliteAAggr is tractable,
	assuming that computation of aggregate functions 
	can be done in time polynomial in the size of the data.
This can be shown by reducing conjunctive query answering
	over ontologies with aggregates to 
	the one over aggregate free ontologies 
	of~\cite{Lutz:2012vh}.
Indeed, consider
	a \dllite ontology $\O'$ and dataset $\D'$ 
	constructed as follows:
	$\O'$ is obtained from $\O$ by replacing all aggregate concepts
	of the form $\circ_r (\agg\ F)$
	with a fresh closed predicate $U$ 
	in every $\O$'s axiom containing $\circ_r (\agg\ F)$;
	$\D'$ extends $\D$ with the set of assertions 
	$\{ F(a, v) \mid  F'(a ,v) \in \D \text{ and } \O \models F' \isa F\}$
	and, 
	$\{ U(a) \mid \agg \bag{d \mid F(a, v) \in \D'}\circ r \}$. 
Observe that $\O'$ is safe according to~\cite{Lutz:2012vh} and, 
	hence, conjunctive query answering is tractable. 
Now, let $Q$ be a conjunctive query	over $\O \cup \D$.
Then, one can easily show that evaluation of a conjunctive query $Q$
	over $\O \cup \D$ gives the same result 
	as evaluation of $Q'$, where each atom of the form 
	$(\circ_r (\agg\ F))(x)$ is replaced with $U(x)$,
	over $\O' \cup \D'$.
Moreover, one can show that 
	the standard 
	query rewriting algorithm of~\cite{Calvanese:2009tf}
	proposed for \dlliteA 
	as a part of query transformation procedure
	(with an extension discussed in
	Section~\ref{sec:mapping-language})
	also works for \dlliteAAggr and SQL.\looseness=-1

\subsubsection{Discussion.}
Note that our aggregate concepts can be encoded as aggregate queries over
attributes as soon as the latter are interpreted under the closed-world
semantics.  
Indeed, given $E = \circ_r(\agg\ F)$,
certain answers for the atomic query $q(x) \text{ :- } E(x)$ 
over this aggregate concept 
would be the same as 
for the following aggregate query over $F$:
\[Q_E(x) = {\sf SELECT} \ \ x \ \ {\sf FROM} \ \ F(x,y) \  \ {\sf GROUP\ BY}\ \ x \ \ {\sf HAVING}\ \ \agg(y) \circ r.\]
Thus, one can reduce conjunctive 
	query answering over our analytics aware \dlliteAAggr
	ontologies to aggregate query answering 
	over classical 
	\dlliteA ontologies as soon as 
	the closed-world semantics is exploited for 
	the interpretation of data attributes.
At the same time, we argue that
	in a number of applications, 
	such as monitoring and diagnostics at 	Siemens~\cite{DBLP:conf/semweb/KharlamovSOZHLRSW14}, 
	explicit aggregate concepts of \dlliteAAggr give 
	us significant modelling and query formulation 	advantages 
	over \dlliteA since
	in such applications concepts are naturally based 
	on aggregate values of potentially many different 	attributes.
For instance, 
	in Siemens the notion of reliability 
	is naturally based on aggregation over 
	various attributes, i.e.,
	it should be modelled as 
	$E_i \isa \mathit{Reliable}$ for many dfferent 
	aggregate concepts $E_i$,
	and reliability is also commonly exploited in
	diagnostic queries.
In the case of \dlliteAAggr,
	in all such diagnostic queries 
	it suffices to use only one atom
	$\mathit{Reliable}(x)$.
In contrast, in the case of \dlliteA,
	each such diagnostic query would 
	have to contain the whole union 
	$\mathit{Reliable}(x) \cup_i Q_{E_i}(x)$.
	(Alternatively, aggregation can be encoded in mappings 
	as discussed in Section~\ref{sec:mapping-language}
	and possibly adresseds with the help of materialised views
	which is a part of our future work---see the end of Section~\ref{sec:conclusion}.)
Thus, Siemens diagnostics queries over \dlliteA
	would be much more complex than the ones over  	\dlliteAAggr.
Moreover, in the case of 
	\dlliteA, $Q_{E_i}(x)$s in 
	such diagnostics queries will have to be adjusted
	each time the notion of reliability is modified,
	while, 
	in the case of \dlliteAAggr, 
	only the ontology and 
	not the queries should be adjusted.

 

\subsection{Query Language}
\label{sec:query-language}

\starql is a query language over ontologies that 
allows to query both streaming and static data 
and supports not only standard aggregates such as \countf, \avgf, etc but also more advanced aggregation functions from our backend system such as Pearson correlation.
In this section
	we will give an overview of the main language 
	constructs and semantics of \starql,
	and illustrate it on our running example
	(see~\cite{oeMoeNeu14streamKI}
	for more details on its semantics).
Each \starql query
	takes as input
	a static \owlaql ontology 
	and dataset
	as well as a set of live and historic streams.
The output of the query is 
	a stream of timestamped data assertions
	about objects
	that occur in the static input data and
	satisfy two kinds of filters:
	\begin{inparaenum}[\it (i)]
		\item 
		a conjunctive query over the input static ontology and data
		and 
	\item 
		a diagnostic query over the input streaming data---which can be live and archived 
		(i.e., static)---
		that may involve typical mathematical, 
		statistical, 
		and event pattern features needed in real-time
		diagnostic scenarios.		
	\end{inparaenum}
The syntax of \starql  
	is inspired by the W3C standardised 
	SPARQL query language; %
it also allows for nesting of queries.
Moreover, \starql has a formal semantics that
	combines open and closed-world reasoning
	and extends snapshot semantics for window 
	operators~\cite{arasu2006cql}
	with sequencing semantics 
	that 
	can handle integrity constraints such 
	as functionality assertions.

In Fig.~\ref{fig:starqlExample}
	we present a \starql query 
	that captures the diagnostic task 
	from our running example
	and uses concepts, roles, and attributes from
	our Siemens
	ontology~\cite{DBLP:conf/semweb/KharlamovSOZHLRSW14,DBLP:conf/debs/KharlamovBGJKLM16,Kharlamov:2016:SIGMOD-demo,DBLP:conf/semweb/KharlamovBGJLNO15,DBLP:conf/semweb/KharlamovJPRSSX15,DBLP:conf/esws/KharlamovJZBGHHKKORRSSSW13,Kharlamov:2013vo}
	and Eq.~\eqref{eq:ontology}.
The query has three  parts:
	declaration of the output stream (Lines~5 and~6),
	sub-query over the static data (Lines~8 and~9)
	that in the running example corresponds to 
	`\emph{return all temperature sensors that are reliable, i.e., with the average score of validation tests at least 90\%}'
	and
	sub-query over the streaming data (Lines~11--17)
	that in the running example corresponds to
	`\emph{whose measurements within the last 10 min Pearson correlate by at least 0.75 to measurements reported by a reference sensor last year}'.	
Moreover, in Line~1  
	there is declarations of the namespace that is used 
	in the sub-queries, i.e., 
	the URI of the Siemens ontology,
	and in Line~3
	there is a declaration of the pulse of the streaming sub-query.
We now enumerate the main clauses 
	of \starql and illustrate them using the query in Fig.~\ref{fig:starqlExample}:
	
\begin{figure}[t!]
\lstdefinestyle{starql}
	{
		keywordstyle=\color{blue},
		basicstyle=\scriptsize\ttfamily,
		language=sql,
		morekeywords={CREATE, STREAM, CONSTRUCT, GRAPH, NOW, SEQUENCE, STATIC, ABOX, TBOX,
		ONTOLOGY, DATA,
		EXISTS, FORALL, MAX, AGGREGATE, IF, THEN, STREAM, PULSE, PREFIX, WITH, START, FREQUENCY, sec},
		numbers=left,
		numberstyle=\tiny,
		numbersep=4pt,
	   	breaklines=true,
		frame={lines},
		rulecolor=\color{gray},
		framerule=0.5pt,
		backgroundcolor=\color{lightgray!15},
		aboveskip=1em,
		belowskip=1em,
		showstringspaces=false,
		tabsize=3,
		framesep=1em	
	}	
\begin{lstlisting}[style=starql]
PREFIX ex : <http://www.siemens.com/onto/gasturbine/>

CREATE PULSE examplePulse WITH START = NOW, FREQUENCY = 1min

CREATE STREAM StreamOfSensorsInCriticalMode AS
CONSTRUCT GRAPH NOW { ?sensor a :InCriticalMode }

FROM STATIC ONTOLOGY ex:sensorOntology, DATA ex:sensorStaticData
WHERE { ?sensor a ex:Reliable } 

FROM STREAM   sensorMeasurements 			[NOW - 1min, NOW]-> 1sec
				  referenceSensorMeasurements 1year <-[NOW - 1min, NOW]-> 1sec,	
USING PULSE   examplePulse
SEQUENCE BY   StandardSequencing AS MergedSequenceOfMeasurementes 
HAVING EXISTS i IN MergedSequenceOfMeasurementes 
		(GRAPH i { ?sensor ex:hasValue ?y. ex:refSensor ex:hasValue ?z }) 
		HAVING PearsonCorrelation(?y, ?z) > 0.75
\end{lstlisting}
\caption{Running example query expressed in \starql}
\label{fig:starqlExample}
\end{figure}

\begin{enumerate}
	\item[{\tt CREATE STREAM}] 
	 clause declares the name of the output stream.
	In our example the output stream is called
	\emph{StreamOfSensorsInCriticalMode}.
	
	\item[{\tt SELECT/CONSTRUCT}]  
	clause defines how the output stream declared 
	in the previous clause should be formed. 
	\starql allows for two types of output:
	the {\tt SELECT} clause forms the output
	as simply the lists of variable bindings,	
	and the {\tt CONSTRUCT} clause
	defines the output as an RDF graph 
	that further can be stored in an RDF datastore 
	or sent as input to another \starql query.
	In our example, 
	we form the output as 
	a set of data assertion of the form $A(b)$,
	thus making an RDF graph consisting of
	all sensors (i.e., {\tt ?sensor}) that function 
	in a critical mode (i.e, {\tt ex:InCriticalMode}) and are 
	determined by the two sub-queries.

	\item[{\tt FROM STATIC/STREAM}]  
	clause declares input static ontology and data 
	and defines streaming data with window parameters
	using the start and end value,
	e.g., `{\tt [NOW - 1min, NOW]}',
	as well as a slide parameter,
	e.g., `{\tt -> \tt 1sec}'.
	In our example, 
	we have the static ontology 
	{\tt ex:sensorOntology} and data 
	{\tt DATA ex:sensorStaticData}
	and two streams:
	{\tt sensorMeasurements} of live sensor measurements
	and also {\tt referenceSensorMeasurements} of 
	recorded measurements of the reference sensor.
	Note that 
	the recorded sensor  
	uses a set back time of one year,
	that is, values from one year ago are correlated 
	to a live stream.
	
	\item[{\tt USING}]  
	clause defines the periodic pulse for the 
	input streams, 
	given by an execution frequency, e.g., {\tt 1min}
	and its absolute start and/or end time, e.g., {\tt NOW}.
	
	\item[{\tt WHERE}]  
	clause declares a static conjunctive query
	expressed as a SPARQL graph pattern.
	The output variables of this query identify 
	possible answers over the static data.
	In our example,
	the query is $\mathit{Reliable(x)}$ where 
	$x$ corresponds to {\tt ?sensor} 
	in the graph pattern `{\tt ?sensor a ex:Reliable}'.
	
	\item[{\tt SEQUENCE BY}] clause
	defines how the input streams should be merged 
	into one and gives a name to the resulting 
	merged stream.

	\item[{\tt HAVING}] 
	clause declares a streaming query.
	It can contain various constructs, including a conjunctive query
	expressed as a graph pattern,
	applied over all elements of the merged 
	stream that have a specific timestamp identified 
	by an index.
	In our example the query
	`{\tt ?sensor ex:hasValue ?y. ex:refSensor ex:hasValue ?z}'
	which is applied at the index point `{\tt i}' of the merged stream
	and retrieves all measurements values 
	of the candidate sensor (i.e., {\tt ?sensor}) and the reference sensor 
	(i.e., {\tt ex:refSensor}).
	In the {\tt HAVING} clause one can do more than referring to 
	specific timepoints: one can also compare them 
	by evaluating graph patterns on each of the states 
	or just return variables mentioned in the graph pattern, 
	while restricting them by logical conditions 
	or correlations, 
	like the Pearson correlation in our example,
	where we verify that the live values {\tt ?y} of the candidate sensor 
	are Pearson correlated with the archived values {\tt ?z} 
	of the reference sensor.
\end{enumerate}

\starql has more features than what we have described above.
In particular, 
it distinguishes between two kinds of variables that correspond to 
either points of time and their arrangement in the temporal sequence,
or to the actual values defined by graph patterns 
of the {\tt HAVING} or {\tt WHERE} clause. 
Variables of different kinds cannot be mixed
and points in time cannot be part of the output.
Note that the state based relations of the {\tt HAVING} clause
	are safe in the first-order logic sense 
	and can be arranged by filter conditions 
	on the state variables.
This safety condition guarantees 
	{\tt HAVING} clauses are domain 
	independent 
	and thus can be smoothly transformed into
	domain independent queries in the languages of CQL~\cite{arasu2006cql} and \sqls{},
%
which is our extension of SQL for stream handling (see Sec.~\ref{sec:implementation} for more details).

Regarding the semantics of \starql, 
	it combines open and closed-world reasoning and extends snapshot semantics 
	for window 
	operators~\cite{arasu2006cql} with sequencing semantics that can handle integrity constraints such as functionality assertions.
In particular, the window operator in combination with the sequencing operator provides a sequence of datasets on which temporal (state-based) reasoning can be applied.  
Every temporal dataset frequently produced by the window operator is converted to a sequence of (pure) datasets. 
The sequence strategy determines how the timestamped assertions are sequenced into datasets. 
In the case of the presented example in Fig.~\ref{fig:starqlExample}, the chosen sequencing method is \emph{standard sequencing} assertions with the same timestamp are grouped into the same dataset.
So, at every time point, one has a sequence of datasets on which temporal (state-based) reasoning can be applied. 
This is realised in STARQL by a sorted first-order logic template in which state stamped graph patterns are embedded. 
For evaluation of the time sequence, the graph patterns of the static {\tt WHERE} clause are mixed into each state to join static and streamed data. 
Note that \starql uses semantics with a real temporal dimension, 
where time is treated in a non-reified manner as an additional ontological dimension and not as ordinary attribute as, e.g., in SPARQLStream~\cite{DBLP:conf/semweb/CalbimonteCG10}.


\subsection{Mapping Language and Query Transformation}
\label{sec:mapping-language}

In this section we present how ontological \starql queries, $Q_{\sf starql}$,
	are transformed
	into semantically equivalent continuous queries, $Q_{\sf sql^{\varoplus}}$,
	in the language \sqls. 
The latter language is an expressive extension of SQL with the appropriate operators for registering continuous queries against streams and updatable relations.
	The language's operators for handling temporal and streaming information are presented in Sec.~\ref{sec:implementation}.

As schematically illustrated in Eq.~\eqref{eq:rewrite-unfold} below,
	during the transformation process
	the  static conjunctive $Q_{\sf{StatCQ}}$ and
	streaming $Q_{\sf{Stream}}$ parts of $Q_{\sf starql}$,
	are 
	first independently \emph{rewritten} 
	using the `${\sf rewrite}$' procedure
	that relies on the input ontology \O
	into
	the union of static conjunctive queries $Q'_{\sf{StatUCQ}}$
	and  
	a new streaming query $Q'_{\sf{Stream}}$,
	and then \emph{unfolded} 
	using the `${\sf unfold}$' procedure
	that relies on the input mappings \M
	into 
	an aggregate SQL query $Q''_{\sf{AggSQL}}$ and  
	a streaming \sqls query $Q''_{\sf{Stream}}$
	that together give an \sqls query $Q_{\sf sql^{\varoplus}}$,
	i.e.,
	$Q_{\sf sql^{\varoplus}} = {\sf unfold}({\sf rewrite}(Q_{\sf starql}))$:	
	\\[-2ex]
\begin{align}
	\label{eq:rewrite-unfold}
	\nonumber
	Q_{\sf starql} \approx Q_{\sf{StatCQ}} \wedge Q_{\sf{Stream}}
	& \xrightarrow[\O]{\sf rewrite}
	Q'_{\sf{StatUCQ}} \wedge Q'_{\sf{Stream}}
	\\
	& \xrightarrow[\M]{\sf unfold}
	Q''_{\sf{AggSQL}} \wedge Q''_{\sf{Stream}}
	\approx	Q_{\sf sql^{\varoplus}}.\\[-4ex]
	\nonumber
\end{align}
In this process we use the rewriting procedure 
	of~\cite{Calvanese:2009tf}, while 
	the unfolding
	relies on mappings of three kinds:
\begin{inparaenum}[\it (i)]
\item 
\emph{classical}: 
from concepts, roles, and attributes to SQL queries
over relational schemas of static, streaming, or historical data,
\item 
\emph{aggregate}: 
from aggregate concepts to aggregate SQL queries over static data, and
\item 
\emph{streaming}: 
from the constructs of the streaming queries of \starql
into \sqls queries over streaming and historical data.
\end{inparaenum}
Our mapping language extends the one presented 
in~\cite{Calvanese:2009tf} for the classical OBDA setting 
that allows only for the classical mappings.

We now illustrate our mappings 
as well as the whole query transformation procedure.

\subsubsection{Transformation of Static Queries.}
We first show the transformation of the 
example static query 
that asks for reliable sensors.
The rewriting of this query 
with the example ontology axioms from Equation~\eqref{eq:ontology} is the following query:
\[
{\sf rewrite}(\mathit{Reliable}(x)) =
\mathit{Reliable}(x)
\lor (\geq_{0.9}(\min\ \mathit{testScore}))(x).
\]

In order to unfold `${\sf rewrite}(\mathit{Reliable}(x))$'
we need both classical and aggregate mappings.
Consider four classical mappings:
one for the concept `$\mathit{Reliable}$'
and three for the attributes 
`$\mathit{testScore}$' and `$\mathit{precisionScore}$',
where $\sql_i$ are some SQL queries:
\begin{align*}
     \mathit{Reliable}(x) & \leftarrow \sql_1(x),\quad &
	\mathit{testScore}(x,y) & \leftarrow \sql_3(x,y),\\
	\mathit{precisionScore}(x,y) & \leftarrow \sql_2(x,y),      \quad &
	     \mathit{testScore}(x,y) & \leftarrow \sql_4(x,y).
\end{align*}

We define an aggregate mapping for a concept
$E = \circ_r(\agg\ F)$ 
as $E(x) \leftarrow \sql_E(x)$, where $\sql_E(x)$
is an SQL query defined as 
\begin{align}
	\label{eq:aggregate-mapping}
\mbox{}\hspace{-2ex}
\sql_E(x) = {\sf SELECT} \ \ x \ \
{\sf FROM} \ \ {\sf SQL}_F(x,y) \  \
{\sf GROUP\ BY}\ \ x \ \
{\sf HAVING}\ \ \agg(y) \circ r
\end{align}
%
where ${\sf SQL}_F(x,y)={\sf unfold}({\sf rewrite}(F(x,y)))$, i.e., the SQL query obtained as the rewriting and unfolding of the attribute $F$.
Thus, a mapping for our example aggregate concept 
$E = (\geq_{0.9}\sf (min\ \mathit{testScore}))$ 
is 
\begin{align*}
	\sql_{E}(x) =& \mbox{ ${\sf SELECT}$ $x$ ${\sf FROM}$ 
	${\sf SQL}_{\textit testScore}(x,y)$ 
	${\sf GROUP\ BY}$ $x$ 
	${\sf HAVING}$ $\min(y) \geq 0.9$}
\end{align*}
where
	${\sf SQL}_{testScore}(x,y) = 
	\sql_2(x,y)\ {\sf UNION}\ 
	\sql_3(x,y)\ {\sf UNION}\ \sql_4(x,y)$.

Finally, we obtain 
\[
{\sf unfold}({\sf rewrite}(\mathit{Reliable}(x)))
= \sql_1(x)\ \mbox{\sf UNION}\ \sql_{E}(x).
\]

\subsubsection{Discussion.}
Note that one can encode \dlliteAAggr aggregate concepts 
as standard \dlliteA concepts using mappings.
Indeed, 
	one can introduce a new atomic concept $A_E$
	for each concept $E = \circ_r(\agg\ F)$
	and a corresponding mapping 
	$A_E(x) \leftarrow \sql_{E}(x)$,
	where $\sql_{E}(x)$ is as in
	Eq.~\eqref{eq:aggregate-mapping}.
One can show that 
	certain answers to
	the query $Q(x) \text{ :- } E(x)$
	are the same as for the query 
	$Q(x) \text{ :- } A_E(x)$.
We argue, however, that 
	this encoding has practical disadvantages compared to
	our approach with aggregate concepts.
Indeed, 
	in the case of aggregate concepts,
	the SQL query $\sql_E$ 
	that maps $E$ to data 
	is computed on the fly during query transformation
	by `composing' the mapping for 
	the rewritten and unfolded attribute $F$ 
	and the mapping for the 
	`aggregate context' of $F$, $\circ_r(\agg\ \star)$, in $E$.
Thus, $\sql_E$ is not actually stored by the 
	query transformation system as it
	depends on the definition of $F$ in the ontology
	and some relevant mappings
	and may change when the ontology 
	or mappings are modified.
At the same time, 
	if one encodes 
	$E$ with a fresh concept $A_E$ and a mapping
	$A_E(x) \leftarrow \sql_{E}(x)$
	and stores them, 
	then one would have to ensure 
	that each further modification 
	in the ontology and mappings relevant
	to $F$ are propagated in $\sql_{E}(x)$.
Another benefit of using aggregate concepts instead of aggregate queries in mappings 
	is that the former approach offers more modelling flexibility. 
Indeed, consider a data property \emph{HasTemperature}.
One can map it to datasources with potentially many non-aggregate mappings 
	and then a knowledge engineer can define various aggregate concepts 
	required by applications
	(i.e., with \avg or \max temperatures) 
	over this property using only ontological terms.
This approach does not require to write mappings with complex SQL queries
	for each new aggregation required by applications.
Nevertheless,
	both the use of aggregate functions in mappings and in the ontology 
	have their benefits that depend on a concrete application at hand and 
	thus comparison between the two approaches require further investigation.
\looseness=-1

\subsubsection{Transformation of Streaming Queries.}
\label{sec:starqltrans}
The streaming part of a \starql query may involve 
static concepts and roles such as \emph{Rotor}
and \emph{testRotor} 
that are mapped into static data,
and dynamic ones such as \emph{hasValue} that are mapped into streaming data. 
Mappings for the static ontological vocabulary are classical and discussed above.
Mappings for the dynamic vocabulary
are composed from the mappings for attributes 
and the mapping schemata for 
\starql query clauses and constructs.
The mapping schemata rely on user defined functions of \sqls
 and involve windows and sequencing parameters specified in a given \starql query which make them
dependent on time-based relations and temporal states.
Note that the latter kind of mappings is not supported by traditional OBDA systems.
\looseness=-1

	%
	%
	%
%

For instance, 
a  mapping schema for 
the `{\tt GRAPH i}' \starql construct 
(see Line~16, Fig.~\ref{fig:starqlExample})
can be defined based on the following
classical mapping that relates a dynamic attribute $\mathit{ex\hspace{-0.5ex}:\hspace{-0.5ex}hasVal}$ to the table \emph{Msmt} 
about measurements 
that among others has attributes \emph{sid}
and \emph{sval}
for storing sensor IDs and measurement values:
\begin{align*}
\mathit{ex\hspace{-0.5ex}:\hspace{-0.5ex}hasVal} 
(\mathit{Msmt.sid,  Msmt.sval})
\leftarrow 
{\sf SELECT}  \ \mathit{Msmt.sid,  Msmt.sval} \ \
{\sf FROM}  \ \mathit{Msmt}.
\end{align*}
The actual mapping schema
for `{\tt GRAPH i}' extends this mapping as following:
\begin{align*}
{\sf GRAPH} \ \ i  \ \ \{ \mathit{?sensor} \ \ \mathit{ex\hspace{-.5ex}:\hspace{-.5ex}hasVal} \ \ ?y \}   
\leftarrow \ &
{\sf SELECT} \ \ \mathit{sid} \ \ \mathit{as} \ \ \mathit{?sensor}, \ \ 
				\mathit{sval} \ \ \mathit{as} \ \ ?y\\
& {\sf FROM} \ \ {\sf Slice}
			(\mathit{Msmt},i,r,\mathit{sl},\mathit{st}),
\end{align*}
where 
the left part of the schema contains an indexed graph triple pattern 
and the right part extends the mapping
for $\mathit{ex\hspace{-0.5ex}:\hspace{-0.5ex}hasVal}$ by applying a function  
$\mathit{Slice}$ that describes the relevant finite slice of the stream $\mathit{Msmt}$ from which the triples in the $i^{th}$ RDF graph in the sequence are produced
and uses the parameters such as the window range $r$, 
the slide $sl$, the sequencing strategy $st$ 
and the index $i$.  
(See~\cite{Neuenstadt2015} for further details.)

\subsection{Query Optimisation}
\label{sec:query-optimisation}

Since a \starql query consists of
analytical static and streaming parts, 
the result of its transformation 
by the rewrite and unfold procedures
is an analytical data query that 
also consists of two parts
and 
accesses information from both live streams and static data sources.
A special form of static data are archived-streams that, though static in nature, accommodate temporal information that represents the evolution of a stream in time. 
Therefore, our analytical operations can be classified as:
	\begin{inparaenum}[\it (i)]
	%
	\item~\emph{live-stream operations} that refer to analytical tasks involving exclusively live streams; 
	\item~\emph{\mbox{static-data} operations} that refer to analytical tasks involving exclusively static information; 
	\item~\emph{hybrid operations} that refer to analytical tasks involving live-streams and static data that usually originate from archived stream measurements. 
	\end{inparaenum}
For static-data operations 
	we rely on standard database optimisation 
	techniques
	for aggregate functions. 
For live-stream and hybrid operations we 
	developed a number of optimisation
	techniques and execution strategies. 
	
	A straightforward evaluation strategy on complex continuous queries containing \emph{static-data operations} is for the query planner to  compute the static analytical tasks ahead of the live-stream operations.
The result on the static-data analysis will subsequently be used as a filter on the remaining  streaming part of the query.
	A \emph{live-stream optimisation} that has been embedded into our backend system is \emph{adaptive indexing}. 
	Using this technique our system collects statistics during query execution and adaptively decides to build main-memory indexes on batches of cached stream tuples. 
	These indices are used to expedite query processing during a complex operation. 
	For example, when joining two stream sources, we can use the values of the first stream to probe the main-memory indexed windows of the second stream. Such optimisations have a significant impact on low-selectivity joins, since they allow us to skip significant portions of the live stream.

\begin{figure}[t]
\centering
\includegraphics[width=0.95\linewidth]{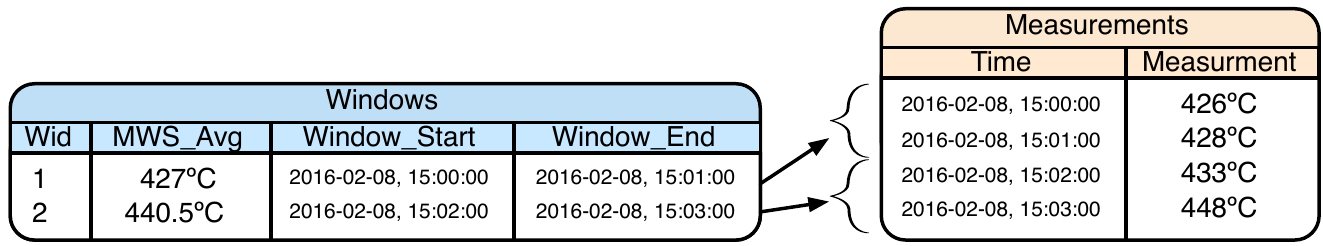}
\caption{Schema for storing archived streams and 
	MWSs
}  
\label{fig: schemata}
\end{figure}

We will now discuss, using an example, the \emph{Materialised Window Signatures} technique for hybrid operations.
Consider the relational schema depicted in Fig.~\ref{fig: schemata} which is adopted for storing archived streams and performing hybrid operations on them. 
	The relational table {\tt Measurements} represents the archived part of the stream and stores the temporal identifier ({\tt Time}) of each measurement and the actual  values (attribute {\tt Measurement}).
	The relational table  {\tt Windows} identifies the windows that have appeared up till now based on the existing window-mechanism. 
	It contains a unique identifier for each window ({\tt Wid}) and the attributes that determine its starting and ending points ({\tt Window\_Start}, {\tt Window\_End}).
	%
The necessary indices that will facilitate the complex analytic computations are materialised. 
	The depicted schema is flexible to query changes since it separates the windowing mechanism ---which is query dependent--- from the actual measurements. 
		
	In order to accelerate analytical tasks that include hybrid
        operations over archived streams, we facilitate precompution
        of frequently requested aggregates on each archived window.  		
	We name these precomputed summarisations as  \emph{Materialised Window Signatures} (\mws{\textit{s}}). These MWSs are calculated when past windows are stored in the backend and are later utilised while performing complex calculations between these windows and a 
live stream. 	
	The summarisation values are determined by the analytics under consideration.
	E.g., 
	for the computation of the Pearson correlation, we precompute the \emph{average} value and \emph{standard deviation} on each archived window measurements;
	for the cosine similarity, we precompute the \emph{Euclidean norm} of each archived window;
	for finding the absolute difference between the average values of the current and  the archived windows, we precompute the  \emph{average} value, etc.

        The selected MWSs are stored in the Windows relation with the use of additional columns. 
	In Fig.~\ref{fig: schemata} we see the MWS summary for the $\mathtt{avg}$ aggregate function being included in the relation as an
 attribute termed $\mathtt{MWS\_Avg}$. The application can easily modify the schema of this relation in order to add or drop MWSs, depending on the analytical workload. 

	When performing hybrid operations between the current and archived windows, some analytic operations
	 can be directly computed based on their 
	 MWS
	 values with no need to access the actual archived measurements. This provides significant benefits as it removes the need to perform a costly join operation between the live stream and the, potentially very large, {\tt Measurements} relation.  
		On the opposite, for calculations such as the Pearson correlation coefficient and the cosine similarity measures, we need to perform calculations that require the archived measurements as well, e.g., for computing cross-correlations or inner-products. 
	Nevertheless, the 
	MWS
	approach allows us to avoid recomputing some of the information on each archived window such as its \emph{avg} value and \emph{deviation} for the Pearson correlation coefficient, and  the Euclidean norm of each archived window for the cosine similarity measure. Moreover, in case when there is a selective additional filter on the query (such as the avg value exceeds a threshold), by creating an index on the $\mathtt{MWS}$ attributes, we can often exclude large portions of the archived measurements from consideration, 
 by taking advantage of the underlying index. 
	%
		%


\section{Implementation}
\label{sec:implementation}
\begin{figure}[t!]
\centering
\includegraphics[width = 1\textwidth]{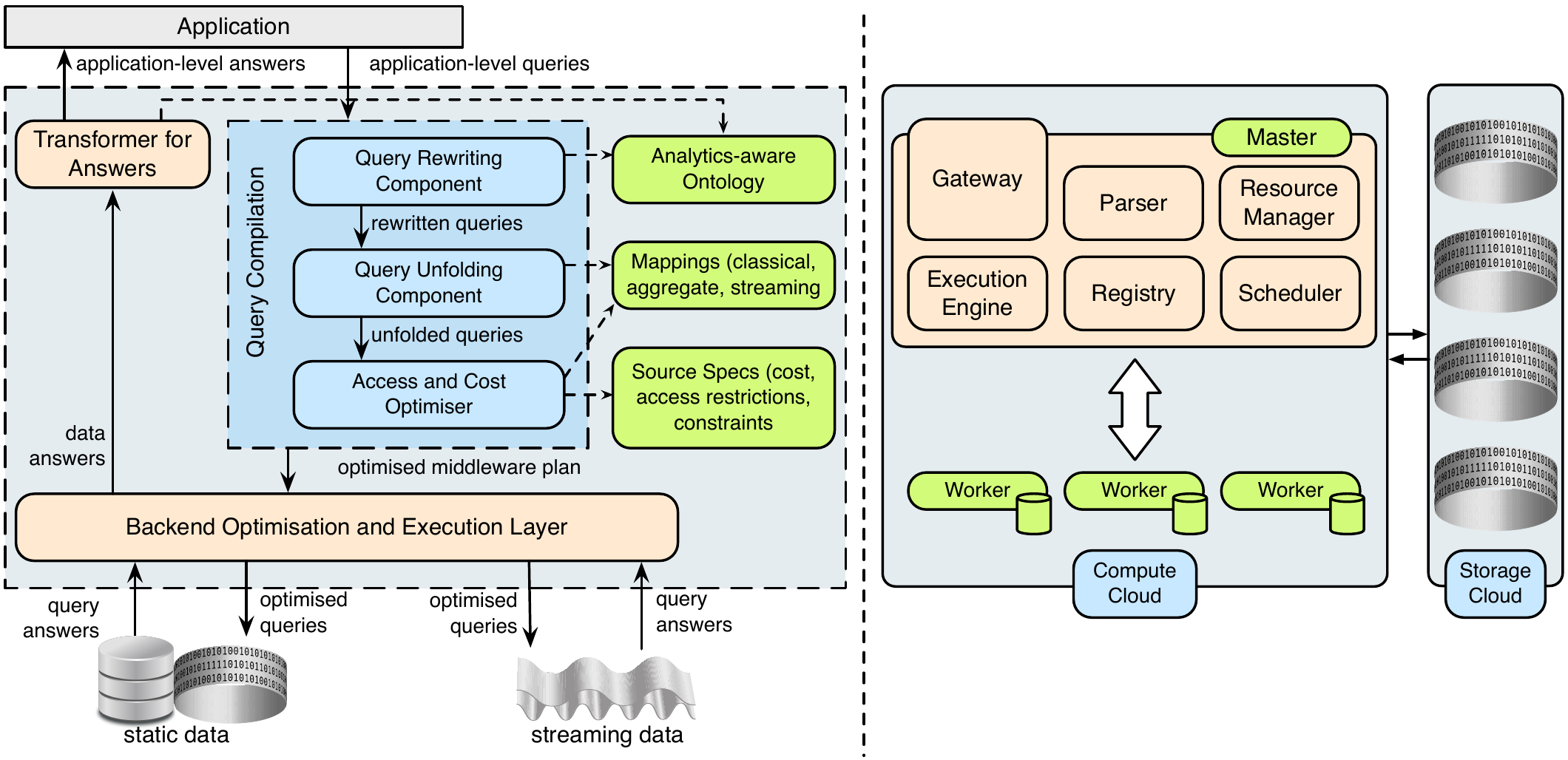}	
\caption{(Left) General architecture. (Right) Distributed stream engine of \exastream}
\label{fig:general architecture}
\end{figure}

In this section we discuss our system that
implements the OBDA extensions proposed in Sec.~\ref{sec:analytics-aware-obda}.
In Fig.~\ref{fig:general architecture} (Left),
we present a general architecture of our system.
On the application level one can formulate 
\starql queries over analytics-aware ontologies
and pass them to the query compilation module
that performs query rewriting, unfolding, and 
optimisation.
Query compilation components can access relevant information in the ontology for query rewriting,
mappings for query unfolding, 
and source specifications for optimisation of
data queries.
Compiled data queries are sent to a query execution
layer that performs distributed query evaluation 
over streaming and static data,
post-processes query answers, and sends them back to applications.
In the following we will discuss two main 
components of the system, namely, 
our dedicated \starqlcql translator that turns \starql queries to \sqls queries,
and our native data-stream management system \exastream
that is in charge of data query optimisation and
distributed query evaluation.
\looseness=-1


\subsubsection{\starql to \sqls Translator.}
Our translator consists of several modules for transformation of various query components and we now give some highlights on how it works. 
The translator starts by turning 
the window operator of the input \starql query
and this results in 
a \emph{slidingWindowView} on the backend system 
that consists of columns for defining \emph{windowID}
(as in Fig.~\ref{fig: schemata}) and \emph{dataGraphID} based on the incoming data tuples.
Our underlying data-stream management system \exastream already provides \emph{user defined functions} (UDFs) 
that automatically create the desired streaming views, e.g., the \emph{timeSlidingWindow} function as discussed below in the \exastream part of the section.

The second important transformation step that we implemented is the transformation of the \starql{} 
{\tt HAVING} clause.
In particular, we normalise the {\tt HAVING} clause into a relational algebra normal form (RANF)
and apply the described slicing technique illustrated in Sec.~\ref{sec:starqltrans}, 
where we unfold each state of the temporal sequence into slices of the \emph{slidingWindowView}.
For the rewriting and unfolding of each slice, we make use of available tools using the OBDA paradigm in the static case, i.e., the Ontop framework~\cite{RodriguezMuro:2013cs}.
After unfolding, we join all states together based on their temporal relations given in the {\tt HAVING} sequence. 
\looseness=-1

\subsubsection{\exastream Data-Stream Management System.}
Data queries produced by the \starqlcql translation, are handled by \exastream which is embedded in
	\exareme, 
	a system for elastic large-scale dataflow processing in the cloud~\cite{tsangaris2009dataflow,kllapi2015elastic}. 
	
\exastream is built as a streaming extension of the
	SQLite database engine, taking advantage of existing Database Management
	technologies and optimisations. 
	It provides the declarative language \sqls for querying data streams and relations.  
	\sqls extends SQL with \udf{s} that incorporate the algorithmic logic for transforming SQLite into a \emph{Data Stream Management Systems} (\dsms). 
	E.g., the \emph{timeSlidingWindow} operator groups tuples from the same time window and associates them with a unique window id.
	In contrast to
	other \dsms{s}, 
	the user does not need to consider low-level
	details of query execution.
	Instead, the system's \emph{query planner} is responsible for
	choosing an optimal plan depending on the query, the available
	stream\slash static data sources, and the execution
	environment.  

\exastream system exploits parallelism in order to accelerate the process of analytical tasks over thousands of stream and static sources. 
	It manages an elastic cloud infrastructure and  dynamically distributes queries and data (including both streams and static tables) to multiple worker nodes that process them in parallel.
The architecture of \exastream's
distributed stream engine is presented in Fig.~\ref{fig:general architecture} (Right).
One can see that queries are registered through the Asynchronous Gateway Server.  
	Each registered query passes through the \exastream parser and then is fed to the  Scheduler module.  
	The Scheduler places the stream and relational operators on worker nodes based on the node's load. 
	These operators are executed by a Stream Engine instance running on each node.

\section{Evaluation}
\label{sec:evaluation}

The aim of our evaluation is to study how
the 
MWS
technique and query distribution to multiple workers
accelerate the overall execution time of analytic queries that correlate a live stream with multiple archived stream records.  

\subsubsection{Evaluation Setting.}
We deployed our system to the Okeanos Cloud         Infrastructure (\url{www.okeanos.grnet.gr/})
and used up to $16$ virtual machines (VMs) each having a
2.66 GHz processor with 4GB of main memory.
	We used streaming and static data  that contains measurements produced by $100,000$ thermocouple sensors installed in 950 Siemens power generating turbines.  
	For our experiments, we used three \emph{test queries} calculating the similarity between the current live stream window and 100,000 archived ones. 
	In each of the test queries we fixed the window size to 1 hour which corresponds to 60 tuples of measurements per window. 
The first query is based on the one 
from our running example (see Fig.~\ref{fig:starqlExample})
which we modified so that it can correlate a live stream with a varying number of archived streams.  
Recall that this query evaluates window measurements similarity 
based on the Pearson correlation. 
The other two queries are variations of the first one where, instead of the Pearson correlation, 
they compute similarity based on 
either the \emph{average} or the \emph{minimum} values 
within a window. 
We defined such similarities between vectors (of measurements) $\vec{w}$ and $\vec{v}$ as follows: \mbox{$|\text{avg}(\vec{w})-\text{avg}(\vec{v})|<
          10^{\circ} C$} 
and $|\text{min}(\vec{w})-\text{min}(\vec{v})|< 10^{\circ} C$.
The archived streams windows are stored in the {\tt Measurements} relation, against which the current stream is compared.

\subsubsection{MWS Optimisation.}
\begin{figure}[t]
\hspace{-3.5ex}
\includegraphics[width=1.04\linewidth]{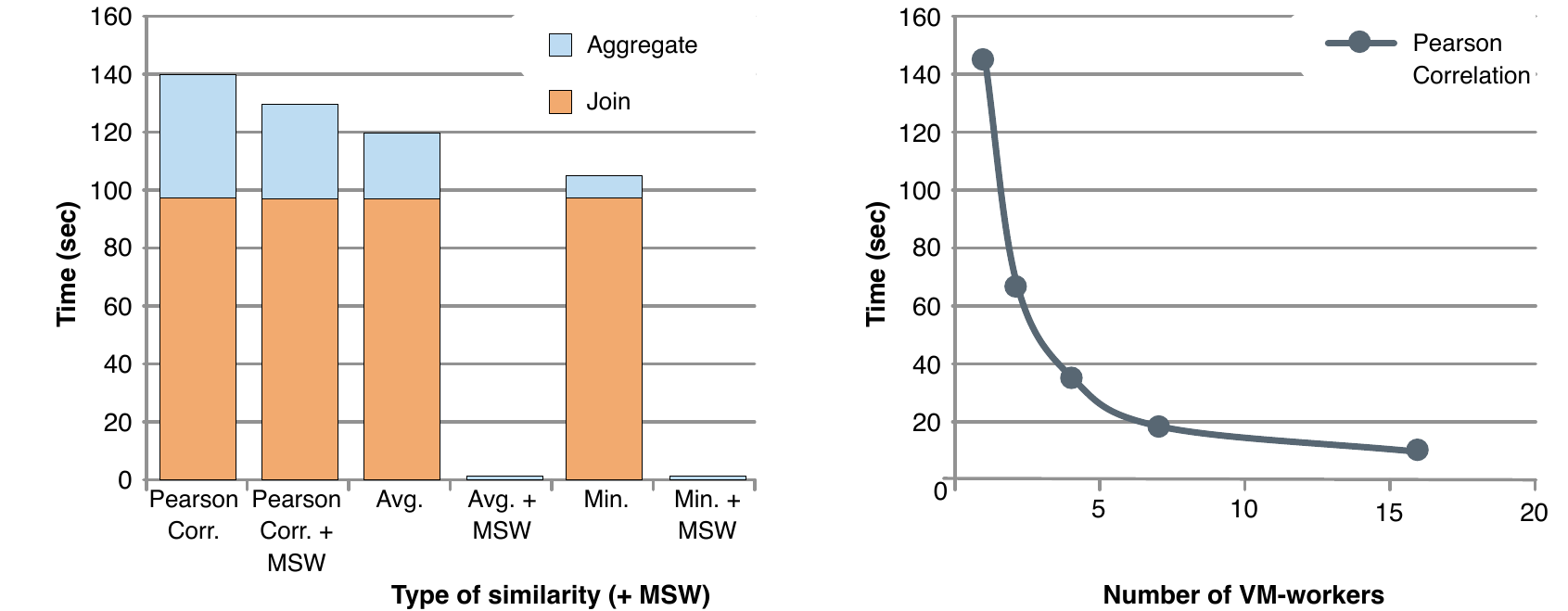}
	\caption{
		(Left) Effect of MWS optimisation
		(Right) Effect of intra-query parallelism}
\label{fig:experiments}
\end{figure}

This set of experiments is devised to show how the 
MWS
optimisation affects the query's response time. 
We executed each of the three test queries on a single VM-worker with and without the 
MWS
optimisation. 
In Fig.~\ref{fig:experiments} (Left) we present the results 
of our experiments. 
The reported time is the average of 15 consecutive live-stream execution cycles.
The horizontal axis displays the three test queries with and without the 
MWS
optimisation, while
the vertical axis measures the time it takes to process 1 live-stream window against all the archived ones. 
This time is divided to the time it takes to join the live stream and the {\tt Measurements} relation and the time it takes to perform the actual computations. 
Observe that the 
MWS
optimisation reduces the time for the Pearson query by 8.18\%. 
This is attributed to the fact that some computations (such as the avg and standard deviation values) are already available in the {\tt Winodws} relation and are, thus, omitted. 
Nevertheless, the join operation between the live stream and the very large {\tt Measurements} relation that takes 69.58\% of the overall query execution time can not be avoided. 
For the other two queries, we not only reduce the CPU overhead of the query, but the optimiser further prunes this join from the query plan as it is no longer necessary. Thus, for these queries, the benefits of the 
MWS
technique are substantial. 
\looseness=-1

\subsubsection{Intra-query Parallelism.}
Since the 
MWS
optimisation substantially accelerates query execution for the two test queries that rely on average and minimum similarities, query distribution would not offer extra benefit, and thus 
these queries were not used in the second experiment.
For complex analytics such as the Pearson correlation that necessitates access to the archived windows, the \exastream backend permits us to accelerate queries by distributing the load among multiple worker nodes.
In the second experiment we use the same setting as before for the Pearson computation without the 
MWS
technique, but we  vary this time the number of available workers from 1 to 16.
In Fig.~\ref{fig:experiments} (Right), one can observe a significant decrease in the overall query execution time as the number of VM-workers increases. 
\exastream distributes the {\tt Measurements} relation between different worker nodes. Each node computes the Pearson coefficient between its subset of archived measurements and the live stream. As the number of archived windows is much greater than the number of available workers, intra-query parallelism results is significant decrease to the time required to perform the join operation. \looseness=-1	
	
To conclude this section, we note that 
MWSs
gave us significant improvements of query execution time for all test queries and parallelism would be essential in the cases where 
MWSs do
not help in avoiding the high cost of query joins since it allows to run the join computation in parallel.
	Due to space limitations, we do not include an experiment examining the query execution times w.r.t. the number of archived windows.
	Nevertheless, based on our observations, scaling up the number of archived windows by a factor of $n$ has about the same effect as scaling down the number of workers by $ \nicefrac{1}{n}$.
	
%
%


\section{Related Work}
\label{sec:related-work}
\subsubsection{OBDA System.} 
Our proposed approach extends existing OBDA systems 
	since they either assume that 
	data is in (static) relational DBs,
	e.g~\cite{DBLP:journals/pvldb/CiviliCGLLLMPRRSS13,RodriguezMuro:2013cs},
	or streaming,
	e.g.,~\cite{DBLP:conf/semweb/CalbimonteCG10,DBLP:conf/semweb/FischerSB13a},
	but not of both kinds.
Moreover, we are different from existing solutions for 
	unified processing of streaming and static semantic data
	e.g.~\cite{DBLP:conf/semweb/PhuocDPH11},
	since they assume that data is natively in RDF 
	while we assume that the data is relational and 
	mapped to RDF.

\subsubsection{Ontology language.} 
The semantic similarities of $\dlliteAAggr$ to other works have been
covered in Sec.~\ref{sec:analytics-aware-obda}. 
Syntactically, the aggregate concepts of $\dlliteAAggr$ have counterpart
concepts, named local range restrictions (denoted by $\forall F.T$) in
$\dlliteA$~\cite{Artale:2012jf}.
However, for purposes of rewritability,
these concepts are not allowed on the left-hand side of inclusion axioms as we
have done for $\dlliteAAggr$, but only in a very restrictive semantic/syntactic
way.
The semantics of $\dlliteAAggr$ for aggregate concepts is very similar to the
epistemic semantics proposed in~\cite{Calvanese:2008hp} for evaluating
conjunctive queries involving aggregate functions. A different semantics based
on minimality has been considered in~\cite{Kostylev:2015dx}. 
Concepts based on aggregates functions were
considered in~\cite{BaaderIS03} 
for languages $\cal ALC$ and $\cal EL$ 
with concrete domains,
but they did not study the problem of query answering.

\subsubsection{Query language.} 
While already several approaches for RDF stream reasoning engines do exist, e.g., CSPARQL~\cite{BARBIERI:2010bw}, RSP-QL~\cite{DellAglio2015} or CQELS~\cite{le2012linked}, 
only one of them supports an ontology based data access approach, namely SPARQLstream~\cite{DBLP:conf/semweb/CalbimonteCG10}.
In comparison to this approach, which also uses a native inclusion of aggregation functions, \starql offers more advanced user defined functions from the backend system like Pearson correlation.

\subsubsection{Data Stream Management System.}
	One of the leading edges in database management systems 
	is to  extend the relational model to support for continuous queries based on  declarative languages analogous to SQL. 
	Following this approach, systems such as  TelegraphCQ~\cite{chandrasekaran2003telegraphcq}, STREAM~\cite{arasu2004stream}, and Aurora~\cite{abadi2003aurora} take advantage of existing Database Management technologies, optimisations, and implementations developed over 30 years of research. 
	In the era of big data and cloud computing, a different class of \dsms has emerged.
	Systems such as Storm 
	and Flink 
	offer an API that allows the user to submit dataflows of user defined operators. 
	\exastream unifies these two different approaches by allowing to describe in a declarative way complex dataflows of (possibly user-defined) operators. 
Moreover, the Materialised Window Signature summarisation, implemented in \exastream,  is inspired from data warehousing techniques for maintaining selected aggregates on stored datasets~\cite{kotidis1999dynamat, gray1997data}. 
We adjusted these technique for complex analytics that blend streaming with static data.


\section{Conclusion, Lessons Learned, and Future Work}
\label{sec:conclusion}

We see our work as a first step towards 
the development of a solid theory 
and new full-fledged systems in the space of
analytics-aware ontology-based access
to data that is stored in different formats 
such as static relational, streaming, etc.
To this end we proposed ontology, query, and mapping languages
that are capable of supporting analytical tasks 
common for Siemens turbine diagnostics.
Moreover, we developed a number of backend optimisation techniques
that allow such tasks to be accomplished in reasonable time
as we have demonstrated on large scale Siemens data.

The lessons we have learned so far are the encouraging evaluation results over the Siemens turbine data (presented in Section~\ref{sec:evaluation}). Since our work is a part of an ongoing project that involves Siemens, we plan to continue implementation and then deployment of our solution in Siemens. 
This will give us an opportunity to do further performance evaluation as well as to conduct user studies.
%


Finally, there is a number of important further research directions 
that we plan to explore.
On the side of analytics-aware ontologies, 
since bag semantics is natural and important in analytical tasks, we see a need in exploring bag instead of set semantics for ontologies. 
On the side of analytics-aware queries, an important further direction is to align them with the 
terminology of the W3C RDF Data Cube Vocabulary 
and to provide additional optimisations after the alignment.  
As for  query optimisation techniques, exploring approximation algorithms for
fast computation of complex analytics between live and archived streams is particularly important.
That is because these algorithms usually provide quality guarantees about the results and in the average case require much less computation.
Thus, we intend to examine their effectiveness in combination with the 
MWS
approach.
	Another interesting backend optimisation relates to the pre-computation of the appropriate structures that will accelerate the aggregate-query execution, e.g. materialised views and database indexes. 
	We intend to examine refined optimisation techniques that combine information on the \obda layer with building of the appropriate structures on our \dsms (or database engine). 
\looseness=-1


%

 \bibliographystyle{splncs}
 \bibliography{references}
%


 \end{document}